\documentclass[sigconf]{aamas}

\usepackage{balance}
\usepackage{times}
\usepackage{graphicx}
\usepackage{helvet}
\usepackage{courier}
\usepackage{latexsym}
\usepackage{amsmath}
\usepackage{url}
\usepackage[noend]{algorithmic}
\usepackage{algorithm}
\usepackage{mathtools}
\usepackage{enumitem}
\usepackage{color}
\usepackage{microtype}
\usepackage{tabularx}
\usepackage{natbib}
\usepackage{rotating}
\usepackage{soul,color}

\usepackage{amsthm}
\usepackage{subcaption}
\usepackage{bm}

\newtheoremstyle{mytheoremstyle} % name
    {\topsep}                    % Space above
    {\topsep}                    % Space below
    {\itshape}                   % Body font
    {}                           % Indent amount
    {\scshape}                   % Theorem head font
    {.}                          % Punctuation after theorem head
    {.5em}                       % Space after theorem head
    {\thmname{#1}\thmnumber{ #2}\thmnote{ (#3)}}  % Theorem head spec (can be left empty, meaning ‘normal’)

\newcommand{\R}{\mathbb{R}}

\theoremstyle{mytheoremstyle}

\setcopyright{ifaamas}
\acmConference[AAMAS '21]{Proc.\@ of the 20th International Conference on Autonomous Agents and Multiagent Systems (AAMAS 2021)}{May 3--7, 2021}{London, UK}{U.~Endriss, A.~Now\'{e}, F.~Dignum, A.~Lomuscio (eds.)}
\copyrightyear{2021}
\acmYear{2021}
\acmDOI{}
\acmPrice{}
\acmISBN{}

\renewcommand\footnotetextcopyrightpermission[1]{} % removes footnote with conference information in first column
\title{Cooperative-Competitive Reinforcement Learning with History-Dependent Rewards}

\author{Keyang He}
\affiliation{
  \department{Department of Computer Science}
  \institution{University of Georgia}}
\email{keyang@uga.edu}

\author{Bikramjit Banerjee}
\affiliation{
  \department{Department of Computer Science}
  \institution{University of Southern Mississippi}}
\email{Bikramjit.Banerjee@usm.edu}

\author{Prashant Doshi}
\affiliation{
  \department{Department of Computer Science}
  \institution{University of Georgia}}
\email{pdoshi@cs.uga.edu}

\begin{document}

\pagestyle{fancy}
\fancyhead{}

\begin{abstract}
Consider a typical organization whose worker agents seek to collectively cooperate for its general betterment. However, each individual agent simultaneously seeks to act to secure a larger chunk than its co-workers of the annual increment in compensation, which usually comes from a {\em fixed} pot. As such, the individual agent in the organization must cooperate and compete. Another feature of many organizations is that a worker receives a bonus, which is often a fraction of previous year's total profit. As such, the agent derives a reward that is also partly dependent on historical performance. How should the individual agent decide to act in this context? Few methods for the mixed cooperative-competitive setting have been presented in recent years, but these are challenged by problem domains whose reward functions do not depend on the current state and action only. Recent deep multi-agent reinforcement learning (MARL) methods using long short-term memory (LSTM) may be used, but these adopt a joint perspective to the interaction or require explicit exchange of information among the agents to promote cooperation, which may not be possible under competition. In this paper, we first show that the agent's decision-making problem can be modeled as an interactive partially observable Markov decision process (I-POMDP) that captures the dynamic of a history-dependent reward. We present an {\em interactive} advantage actor-critic method (IA2C$^+$), which combines the independent advantage actor-critic network with a belief filter that maintains a belief distribution over other agents’ models. Empirical results show that IA2C$^+$ learns the optimal policy faster and more robustly than several other baselines including one that uses a LSTM, even when attributed models are incorrect.\footnote{Preprint. Under review.}
%Additionally, IA2C retains consistency at greater levels of noise in the learners’ observations than the baselines.
\end{abstract}


\maketitle

%-------------------------------------------------------------------------
\section{Introduction}
\label{sec:intro}
%-------------------------------------------------------------------------

Real-world multi-agent domains involving interactions among agents are often not purely cooperative or competitive. For example, consider a typical organization whose worker agents seek to collectively cooperate for its general betterment. However, each individual agent simultaneously also seeks to act to secure a larger chunk than its co-workers of the annual increment in compensation, which usually comes from a {\em fixed} pot. Another feature of many organizations is that a worker receives a bonus, which is often a fraction of the previous year's total profit. Thus, the agent derives a reward that partly depends on historical performance. The individual agent in an organization then faces a context where it must cooperate and compete while collecting history-dependent rewards.

A potential approach to solving the worker agent's decision-making problem is multi-agent reinforcement  learning (MARL). Interest in MARL has grown dramatically over the past decade. 
While traditional MARL has made significant strides in fields such
as  game  playing (e.g.,  AlphaStar~\cite{Vinyals19:Grandmaster})  and
robotics,  it  struggles  in  problems  involving  interactions  where
the individual  and  group interests  may  conflict  with each  other. 
In response    to   this    challenge,    MARL suited to both 
cooperative and competitive  settings  has  received attention recently~\cite{maddpg,coma,lola}.   
However, most of these  recent methods take a joint perspective to the interaction and require explicit exchange of information among the learning agents, which may not be possible under competition.

In this paper, we first introduce and formalize the {\sf Organization} problem as a quintessential domain involving mixed cooperation-competition, while exhibiting both partial observability and history-dependent rewards. Next, we show that the individual agent's decision making can be modeled using an interactive partially observable Markov decision process (I-POMDP)~\cite{Gmytrasiewicz2005}  despite the presence of history-dependent rewards in the problem, and introduce an approach to MARL for this type of problems. In particular, we introduce an interactive advantage actor-critic (labeled IA2C$^+$) algorithm, which combines an independent advantage actor-critic~\cite{a2c}  with a belief   filter  that  maintains  a  belief distribution over the other agents' models, and updates the belief using private observations. We show that even when the set of models attributed to the other agents may not contain a grain of truth, agents using IA2C$^+$ still converge to the optimal policy significantly faster than several relevant baselines, and remain consistent for greater levels of noise in their observations, compared to those baselines. 

%-------------------------------------------------------------------------
\section{Background}
\label{sec:background}
%-------------------------------------------------------------------------
In this section, we briefly review the two main components on which this work is built: a well-known model of decision making in a multi-agent environment, and the actor-critic~\cite{actor-critic} reinforcement learning (RL) ~\cite{Sutton98:Reinforcement} approach used in our solution.

\subsection{Overview of Interactive POMDPs}
\label{sec:ipomdp}

Interactive partially observable Markov decision processes  are a generalization of POMDPs~\cite{pomdp} to sequential decision-making in multi-agent environments~\citep{Gmytrasiewicz2005,Doshi12:Decision}. Formally, an I-POMDP for agent $i$ in an environment with one other agent $j$ is defined as,
\[
\text{I-POMDP}_i = \langle IS_i, A, T_i, O_i, Z_i, R_i, OC_i \rangle
\]

\noindent $\bullet~IS_i$ denotes the interactive state space. This includes the physical state $S$ as well as models of the other agent $M_j$, which may be intentional (ascribing beliefs, capabilities and preferences) or subintentional~\citep{dennett1986intentional}. Examples of the latter are probability distributions and finite state machines. In this paper, we ascribe subintentional models to the other agent.\\
$\bullet~A = A_i \times A_j$ is the set of joint actions of both agents.\\
$\bullet~T_i$ represents the transition function, $T_i$: $S \times A \times S \xrightarrow{} [0, 1]$. The transition function is defined over the physical states and excludes the other agent's models. This is a consequence of the model non-manipulability assumption, which states that an agent's actions do not directly influence the other agent's models.\\
$\bullet~O_i$ is the set of agent $i$'s observations.\\ 
$\bullet~Z_i$ is the observation function, $Z_i$: $A \times S \times O \xrightarrow{} [0, 1]$. The observation function is defined over the physical state space only as a consequence of the model non-observability assumption, which states that other's model parameters may not be observed directly.\\
$\bullet~R_i$ defines the reward function for agent $i$, $R_i$: $S \times A \xrightarrow{} \R$. The reward function for I-POMDPs usually assigns preferences over the physical states and actions only.\\
$\bullet~OC_i$ is the subject agent's optimality criterion, which may be a finite horizon $H$ or a discounted infinite horizon where the discount factor $\gamma \in (0,1)$. 

Without loss of generality, let the subintentional model ascribed to $j$ take the form $m_j = (h_j, \pi_j, Z_j)$ where $h_j$ is agent $j$'s action-observation history, $\pi_j$ is a candidate policy, and $Z_j$ is its observation function. Given agent $i$'s belief over interactive states $b_i$, on action $a_i$ and receiving observation $o_i$, the belief is updated as:
\begin{align}
    b_i'(is') &\propto \sum_{is}b_i(is)\sum_{a_j}Pr(a_j|m_j)Z_i(a,s',o_i')T_i(s,a,s')\nonumber\\
    &\times \sum_{o_j'}\delta_K(APPEND(h_j,o_j'), h_j')Z_j(s',a,o_j')
    \label{eqn:bu}
\end{align}
where $\alpha$ is a normalizing constant, $h_j$ and $h_j'$ are part of $m_j$ and $m_j'$, respectively. $\delta_K$ is the Kronecker-delta function, and APPEND returns a string with the second argument appended to the first.

Each belief state $b_i$ is associated with a value given by:
\begin{align}
    V(b_i) = &\max_{a_i \in A_i}\{\sum_{is}\sum_{a_j}R_i(s,a_i,a_j)Pr(a_j|m_j)b_i(is)\nonumber\\
    & + \gamma\sum_{o_i \in O_i} Pr(o_i|a_i, b_i)V(b_i'(is'))\}
\end{align}
where $b_i'(is')$ is obtained as shown in equation~\ref{eqn:bu}.

\subsection{Actor-Critic RL}
\label{subsec:AC}

RL problems are typically modeled using {\em  Markov decision
  processes} or  MDPs~\cite{Sutton98:Reinforcement}, which  is defined
by  the  tuple $\langle  S,A,T,R\rangle$,  where  $S$  is the  set  of
{\em perfectly observed}  environmental states; $A$ is the set of agent's actions; 

$T(s,a,s')$ is the state transition probability
function specifying  the probability  of the  next
state in the  Markov chain being $s'$ on the agent selecting
action  $a$ in  state $s$;
$R(s,a)$  is the  reward
function  specifying the  reward from  the environment
that the agent  gets for executing action $a\in A$  in state $s\in S$.
 The  agent's goal is  to learn  a policy
$\pi:S\mapsto A$ that maximizes the  sum of current and future rewards
from any state $s$, given by,
\begin{equation*}
V^{\pi}(s)=\mathbb{E}_{T}[R(s,\pi(s))+\gamma R(s',\pi(s'))+\gamma^2\ldots]
\end{equation*}
where $s,s',\ldots$ are successive samplings from the distribution $T$ following the Markov chain with policy $\pi$, and $\gamma\in(0,1)$
is a discount factor. This is sometimes facilitated by learning an action value function, $Q(s,a)$ given by 
\begin{equation}
Q(s,a)=R(s,a)+\max_{\pi}\gamma \sum_{s'}T(s,a,s')V^{\pi}(s').
\label{eqn:Q}
\end{equation}

There are  two main categories of  reinforcement learning: value-based
and  policy-based RL.  On the  one hand,  value-based RL  learns an
optimal value  function (e.g., $V$  given above) that maps  each state
(or state-action  pair) to a  value. It  is more sample  efficient and
stable compared to  policy-based RL, but it usually  requires that the
action space be finite. On the  other hand, policy-based RL learns the
optimal policy directly, sometimes without  using a value function. It
is useful  when the action space  is continuous or stochastic,  and it
has a faster convergence due to directly searching the policy space.

Actor-critic methods take advantage of both  value-based 
policy-based RL while eliminating some drawbacks. It splits the model  into two components, an actor and a
critic, where  the actor controls how  the agent acts by  learning the
optimal  policy,  and the  critic  evaluates  the actor's  actions  by
computing  the  action  value   ($Q$  value  in  equation~\ref{eqn:Q})  function, or directly the value function ($V$). 
The  actor and the critic are optimized  separately during  the  training.   By having  them
interact  with and  complement  each other,  the architecture  is  more  robust  than  if  the two  models  were  used individually.

%\subsection{Advantage Actor-Critic}

Value-based RL  methods typically display  a high variance due  to the
uncertainty  embedded  in the  agent's  experience.  To mitigate  this,  instead of  using the $Q$ value  for the  critic, advantage actor-critic (A2C)  uses {\em advantage} values, given for a policy $\pi$ by 
\begin{align*}
A(s, a) &= Q(s, a) - V^\pi(s)\\
&=R(s,a) + \gamma \sum_{s'}T(s,a,s')V^\pi(s') - V^\pi(s)
\end{align*}
Advantage represents how much better a particular action is at a state compared to the value of the state. While the critic can be optimized by reducing the mean square of the advantages estimated from samples, the actor, $\pi_{\bm \theta}$ (parametrized by ${\bm \theta}$) can be optimized by gradient descent using gradients:
\begin{equation*}
    \mathbb{E}_{s\sim d^{\pi_{\bm \theta}},a\sim\pi_{\bm \theta}}\nabla_{\bm \theta}\log\pi_{\bm \theta}(a|s)A(s,a),
\end{equation*}
where $d^{\pi_{\bm \theta}}(s)=\sum_{t=0}^{\infty}\gamma^tPr(s_t=s|s_0,\pi_{\bm \theta})$ is the discounted state distribution that results from following policy $\pi_{\bm \theta}$.

In an I-POMDP setting, the state is not directly observable; instead the learner receives an observation that is usually (noisily) correlated with the (hidden) state and other agents' actions. The advantage function can be reformulated in terms of belief as 
\begin{align}
A(b_i,a_i)&=\sum_{is}\{\sum_{a_j}R_i(s,a_i,a_j)Pr(a_j|m_j)b_i(is)\nonumber \\ &+\gamma\sum_{o_i}Pr(o_i|a_i,b_i)V^\pi(b_i'(is'))-V^\pi(b_i(is))\}
\label{eqn:advantage}
\end{align}
Assuming $\pi_{\bm\theta}$ maps beliefs to actions, its gradients are
\begin{align}
\mathbb{E}_{b\sim b^{\pi_{\bm\theta}},a\sim\pi_{\bm\theta}}\nabla_{\bm \theta}\log\pi_{\bm \theta}(a|b)A(b,a).
\label{eqn:gradient}
\end{align}
where $b^{\pi_{\bm \theta}}(is)=\sum_{t=0}^{\infty}\gamma^tPr(is_t=is|is_0,\pi_{\bm \theta})$ is the discounted belief distribution that results from following policy $\pi_{\bm \theta}$.

%-------------------------------------------------------------------------
\section{The Organization Domain}
\label{sec:domain}
%-------------------------------------------------------------------------
We introduce the new {\sf Organization} domain, which models a typical business
organization featuring a mix of cooperation toward the overall improvement of the organization and individual competition. Notably, the reward function may not have the Markovian property. To be more specific, a proportion of the rewards from the past is added to the current reward as a bonus to the worker. For
example, if the  organization operated well in the past  year, then it
earned not only a profit from  its business but also reputation about
its business. 
Thus, the measure of its current reward must accommodate this carried-over influence from past rewards. We  call this  a  {\em  history-dependent}  reward.   The goal of each agent is  to maximize the sum of all reward signals that it receives, which includes this history-dependent reward.

\subsection{Specification}

\subsubsection{State and Observation Sets}

\begin{figure}
\includegraphics[width=3in]{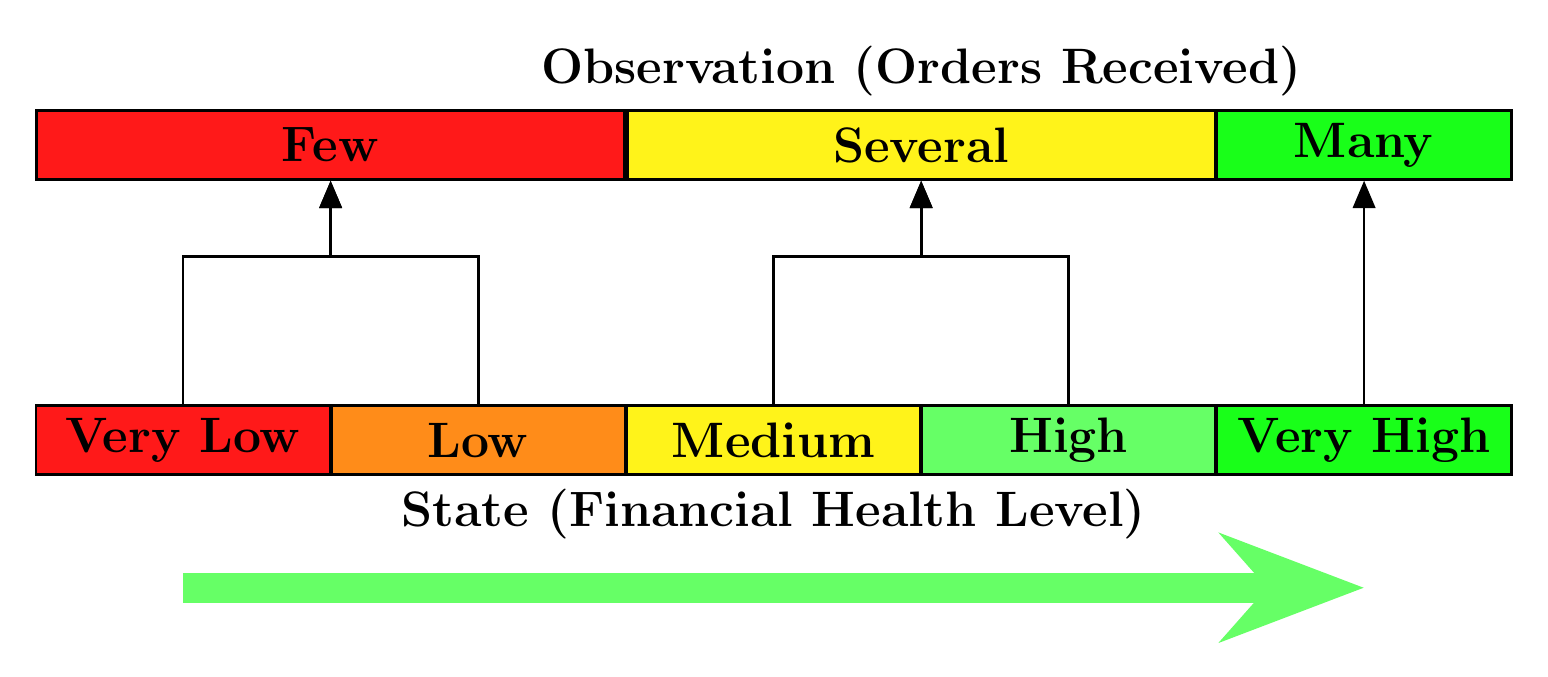}
\caption{\small States of  the organization domain and  the non-deterministic
  public observations.}
\label{fig:state}
\vspace{-0.05in}
\end{figure}

The state space of the problem represents the organization's financial
health level,  which is discretized  into five states: very  low (denoted as $s_{vl}$), low ($s_l$), medium ($s_m$), high ($s_h$),  and very  high ($s_{vh}$). Agents  cannot  directly
observe the financial health level  of the organization; instead, they
only  receive observations  of the  number of  orders received  by the
organization as  well as  other agents' actions.   We assume  that the
observation is decomposed into  {\em public} and {\em private} observations. While
public  observation  is  common  to  all agents  and  depends  on  the
underlying  state  only,  the  private  observation  represents  other
agents' actions and is perceived by the corresponding  agent only. There
are three  possible public  observations pertaining  to the  number of
orders: $meager$  ($o_e$), $several$  ($o_s$), and $many$  ($o_m$), where
$o_e$ indicates the organization is in either $s_{vl}$ or $s_l$, $o_s$
indicates  the organization  is in  either $s_m$  or $s_h$,  and $o_m$
indicates the organization is in $s_{vh}$.  Three private observations
represent the agents' three  possible actions, respectively.  However,
private observations are  also noisy.  Agents have  0.8 probability of
perceiving the actual  action of the other agent,  and 0.2 probability
of receiving either private observation corresponding to actions which
the other agent did  not take.  Figure~\ref{fig:state} illustrates the
relationship between states and public observations.

\subsubsection{Action Space and Transition Function}
\label{subsec:actions}

\begin{figure}[!ht]
\begin{subfigure}{.48\columnwidth}
    \centering
    \includegraphics[width=\textwidth]{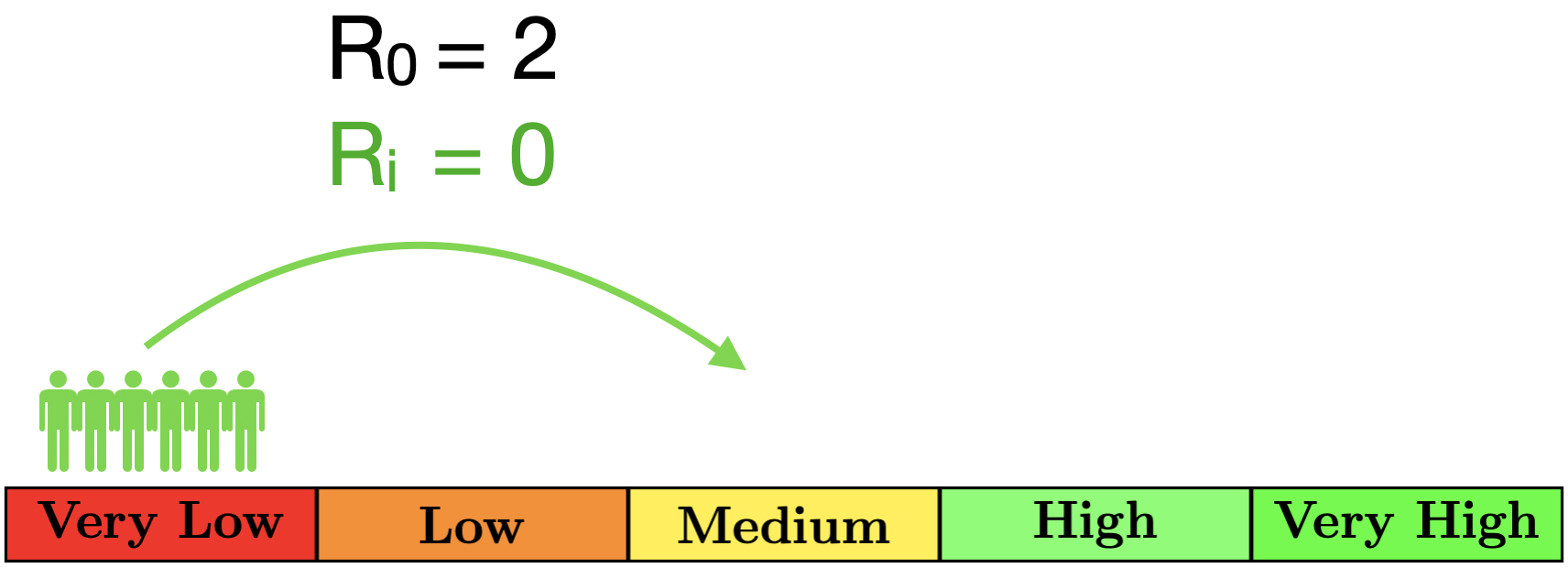}
    \caption{}
    \label{fig:org1}
\end{subfigure}
\begin{subfigure}{.48\columnwidth}
    \centering
    \includegraphics[width=\textwidth]{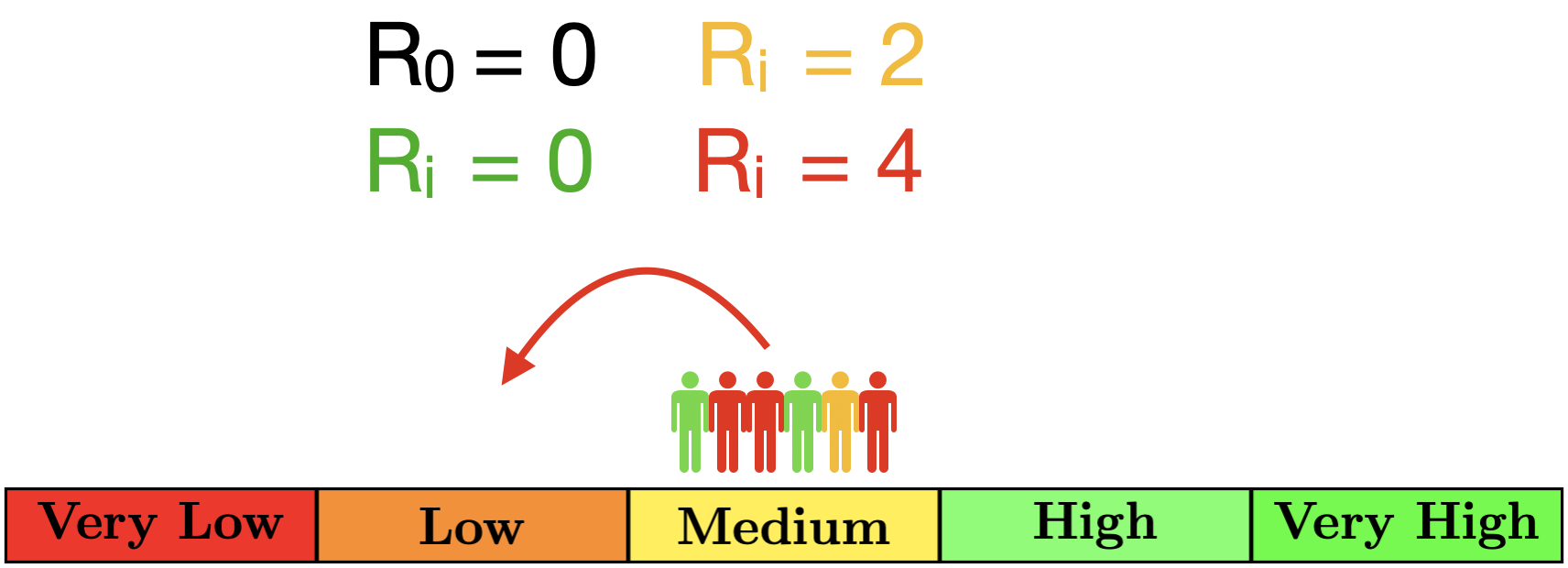}
    \caption{}
    \label{fig:org2}
\end{subfigure}
\begin{subfigure}{.48\columnwidth}
    \centering
    \includegraphics[width=\textwidth]{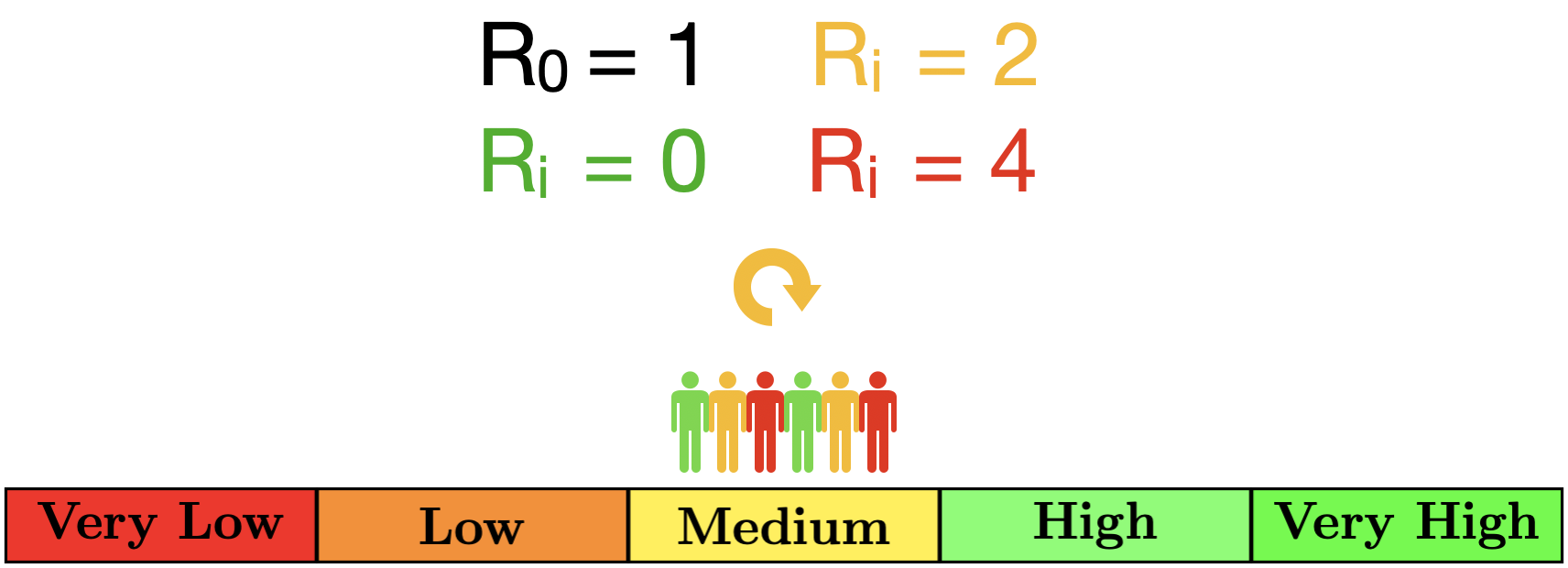}
    \caption{}
    \label{fig:org3}
\end{subfigure}
\begin{subfigure}{.48\columnwidth}
    \centering
    \includegraphics[width=\textwidth]{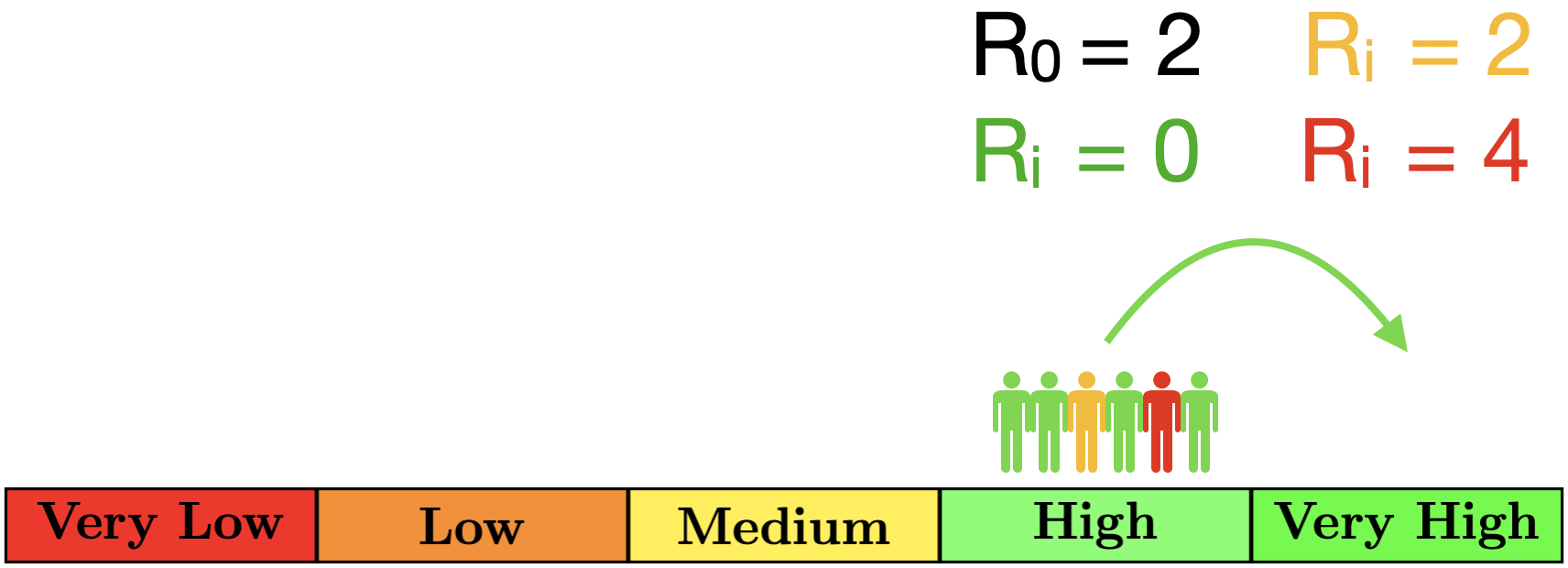}
    \caption{}
    \label{fig:org4}
\end{subfigure}
\caption{\small Agents  picking individual,  balance, and  group actions  are
  colored in red,  yellow, and green, respectively. (a)  If all agents
  pick group  action, the  state increases by  two.  (b)  The majority
  picks individual  action, state decreases  by one. (c) There  is no
  majority action,  state remains  unchanged. (d)  Unanimously picking
  group action  can only increase  the state  by one when  the current
  state is $s_h$; therefore, minority agents may pick other actions to
  receive higher individual rewards.}
\label{fig:org}
\vspace{-0.05in}
\end{figure}

Each  agent  has  three   possible  actions:  $self$,  $balance$,  and
$group$. The  $self$ action  solely benefits the  agent while  the $group$
action benefits the  organization at the expense of  self. The $balance$
action benefits  both self  and the organization.   A joint  action is
determined by the distribution of  the individual agent's actions.  If
the number  of agents who  picked $self$  equals the  number of
agents who  picked $group$, then joint  action is  a $balance$
action.  $Balance$ joint action does  not change the underlying state.
If the  number of agents  who pick $self$  is greater  than the
number of agents  who pick $group$, the joint  action will be a
$self$ action.  $Self$ joint action  brings down the health level (the
current state) by one when the  current state is higher than $s_{vl}$;
otherwise, the state  remains unchanged.  If the number  of agents who
pick  $self$ is smaller  than the  number of  agents who  pick
$group$,  the joint action  will be a $group$  action.  $Group$
joint  action increases  the  current  health level  by  one when  the
current state  is lower  than $s_{vh}$,  otherwise, the  state remains
unchanged.   Besides,  if  all  agents  performed  $group$  individual
action,   the  state   increases   by   two.   Figure   ~\ref{fig:org}
demonstrates state transitions of the organization domain.

\subsubsection{Reward Function}

At time step $t$, each agent receives a sum  of individual and group
rewards, and the bonus.  The agent $i$'s individual reward, $R_i$, depends
on the state and its own action:
$$R_i^t \leftarrow R_i(s^t, a_i^t).$$
The group reward,  common to all agents, depends on  the current state
and joint action:
$$R_0^t \leftarrow R(s^t, a^t).$$
The  bonus or history dependent reward  is a proportion of  the total reward
from the previous time step. For $\phi \in (0,1)$,
$$R_{-1}^t = \phi (\sum\nolimits_i R_i^{t-1} + R_0^{t-1}).$$

The goal of agent $i$ is to optimize the expected sum of individual,
group,               and                history               rewards,
$\mathbb{E}_{trajectories}\left [\sum_t \gamma^t(R_0^t + R_i^t + R_{-1}^t)\right ]$.

The $Group$ action is cooperative giving $R_i = 0$ and $R_0 = r$, where $r \in \R$. The $Self$ action benefits an agent giving reward $R_i = \beta r$ to the  agent, where $\beta > 1$, and $R_0 = 0$.  $Balance$  action gives $R_i = c\frac{1+\beta}{\alpha} r$, where $0<c<1$ and $r < \frac{1+\beta}{\alpha} < \beta$ and  $R_0 = (1-c)\frac{1+\beta}{\alpha} r$, thereby benefiting  both the agent and the business. If the organization  reaches state $s_{vl}$, each  agent receives a penalty of $p$ no  matter which  individual  action  was picked.

\subsection{Importance of History-Dependent Reward}
\label{subsec: history}

In this subsection, we demonstrate the importance of history-\\dependent reward in the {\sf Organization} domain. Suppose that policy $\pi_0$ leads to the reward sequence $\{\beta r, \beta r, r, \dots\}$ for an agent, and policy $\pi_1$ leads to the reward sequence $\{\beta r, \beta r, \frac{1+\beta}{\alpha}r, \frac{1+\beta}{\alpha}r,\dots\}$. For convenience, we set $d = \frac{1+\beta}{\alpha}$. For horizon $H = 4$, the total reward from performing policy $\pi_0$ is:
\begin{small}
\begin{align*}
&\beta r + (\phi \beta r + \beta r) + (\phi^2 \beta r + \phi \beta r + r) + (\phi^3 \beta r + \phi^2 \beta r + \phi r + \beta r)\\
&= \phi^3\beta r + 2\phi^2 \beta r + 2\phi\beta r+ \phi r + 3\beta r + r
\end{align*}
\end{small}
The total reward from performing policy $\pi_1$ is:
\begin{small}
\begin{align*}
&\beta r + (\phi \beta r + \beta r) + (\phi^2 \beta r + \phi \beta r + d r) + (\phi^3 \beta r + \phi^2 \beta r + \phi d r + d r)\\
&= \phi^3\beta r + 2\phi^2 \beta r + 2\phi\beta r + \phi dr + 2\beta r + 2d r
\end{align*}
\end{small}

\vspace{-0.1in}
\begin{figure}[!ht]
    \includegraphics[width=\columnwidth]{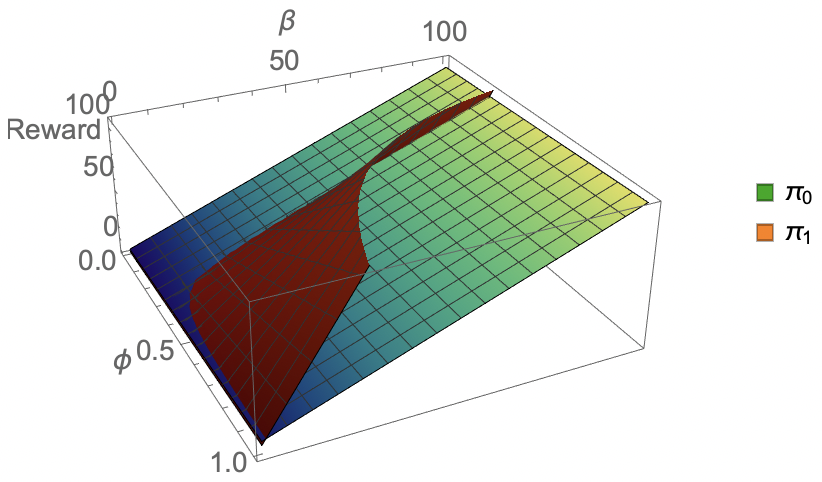}
\caption{\small The total reward from policy $\pi_0$ and $\pi_1$ with varying $\beta$ and $\phi$ for horizon of 4.}
\label{fig:hist}
\end{figure}

Then we can compare the total rewards from these two policies with varying choices of $\beta$ and $\phi$. We use a simple program to check if $\phi$ can affect which of the two policies is optimal for horizon $H \geq 4$ when $d$ is set to $\frac{9}{4}$. The result shows that for every horizon from 4 to 100, the history-dependent parameter $\phi$ is always a deciding factor in finding optimal policy. Figure ~\ref{fig:hist} shows the total reward from policy $\pi_0$ and $\pi_1$ with varying $\beta$ and $\phi$ for horizons 4 and 8. Notice that the parameter $\phi$ influences when each surface has the higher total reward.

\subsection{Modeling the Org Domain as an I-POMDP}
\label{sec:ipomdp-model}

The I-POMDP framework is suitable for modeling the {\sf Organization} domain from an individual agent $i$'s perspective. We formulate such an I-POMDP in this section. Suppose that if we remove the history dependent rewards, then the residual environment has perfectly Markovian dynamics, with functions $T(s_f,a_i, a_j,s_f')$, $Z(a_i, a_j,s_f,s_f',o_f)$, and $R(s_f,a_i, a_j)$, where $s_f$ is an ordinary physical state and $o_f$ is an observation related to this state. (Note that we assume $Z$ depends on both $s_f$ and $s_f'$, unlike in Section~\ref{sec:ipomdp}, as this is needed for our formulation.) To preserve the Markov property even when $R_{-1}$ is introduced, we add a continuous-valued feature, $s_r \in \R$, to the state space, that memorizes the reward from the last step. The new I-POMDP for agent $i$ in the {\sf Organization} domain with one other agent $j$ has an expanded definition: 
\[
\text{I-POMDP}_i = \langle IS_i, A, T_i, \Omega_i, W_i, Z_i, O_i, R_i, OC_i \rangle
\]

\noindent $\bullet$ The interactive state space $IS_i$ now includes the physical state $S_f$, the history-reward state $S_r$, as well as models of the other agent $M_j$. We let the latter be subintentional in this domain.  \\
$\bullet~A = A_i \times A_j$ is the set of joint actions of both agents.\\
$\bullet~T_i$ represents the transition function, now defined as:
\begin{align}
  &T_i(\langle s_f,s_r\rangle, a_i, a_j, \langle s_f',s_r'\rangle) \nonumber\\
  &= \left\{ \begin{array}{ll} T(s_f, a_i, a_j, s_f'), & \text{if }s_r'=R(s_f, a_i, a_j) +\phi\cdot s_r\\
    0, & \text{otherwise}\end{array}\right.
\end{align}

$\bullet~\Omega_i$ is the set of agent $i$'s {\em private} observations.\\ 
$\bullet~W_i: A \times \Omega_i \rightarrow  [0,1]$ is the private observation function.\\
$\bullet~O_i = O_f \times O_r$ is the set of agent $i$'s public observations, where $O_f$ informs about the state and $O_r=S_r$, allowing the agent to observe the past reward.\\
$\bullet~Z_i$ is the observation function, defined as:

\begin{align}
  &Z_i(a_i, a_j, \langle s_f, s_r\rangle, \langle s_f',s_r'\rangle, \langle o_f,o_r\rangle)
  \nonumber\\
  &= \small \left\{ \begin{array}{ll} Z(a_i, a_j, s_f, s_f', o_f), & \text{if }(s_r'=R(s_f,a_i, a_j) +\phi\cdot
    s_r) \land (o_r=s_r')\\\\
    0, & \text{otherwise}\end{array}\right.
\end{align}
$\bullet~R_i$ defines the reward function for agent $i$:
\begin{align}
    R_i(\langle s_f,s_r\rangle,a_i,a_j) = R(s_f, a_i, a_j) + \phi \cdot s_r 
\end{align}
$\bullet OC_i$ is the subject agent's optimality criterion, which may be a finite horizon $H$ or a discounted infinite horizon where the discount factor $\gamma \in (0,1)$.

The belief update equation for the new I-POMDP formulation is:
\begin{align}
&b_i'(is'|b_i,a_i,o_i',\omega_i') = b_i'(\langle s_f',s_r' \rangle|b_i,a_i,o_i',\omega_i')~\times \nonumber\\
&~b_i'(m_j'|\langle s_f', s_r' \rangle, b_i,a_i, o_i', \omega_i')
\label{eq:10}
\end{align}
where the first term can be derived as:
\begin{small}
\begin{align*}
  &b_i'(\langle s_f',s_r' \rangle|b_i,a_i,o_i',\omega_i') \propto \sum_{a_j}\sum_{s_f,s_r=\frac{o_r'-R_i(s_f,a_i, a_j)}{\phi}}Pr(a_j|m_j)\nonumber\\
  &\times T(s_f, a_i, a_j, s_f')b_i(\langle s_f, s_r\rangle)Z(a_i, a_j, s_f, s_f', o_f')\nonumber\\
  &\times Z(a_i, a_j, s_f, s_f', o_f')W_i(a_i,a_j,\omega_i')\delta_k(APPEND(h_j,a_j,o_f'), h_j')
\end{align*}
\end{small}
Recall that $o_f'$ is a public observation received by both agents. We derive the second term of the belief update decomposition in the next subsection. For convenience, we summarize the full update as $\tau(b_i, a_i, o_i', \omega_i', b_i')$.

The Bellman's equation for the new I-POMDP is:
\begin{align}
& V(b_i) = \max_{a_i}\left [\sum_{s_f,s_r}\sum_{a_j}R_i(\langle s_f, s_r\rangle, a_i,a_j) Pr(a_j|m_j)b_i(\langle s_f, s_r\rangle)\right . \nonumber\\
 &+\gamma\sum_{a_j}\sum_{s_f',s_r'=R(s_f,a_i, a_j) +\phi\cdot
    s_r}Pr(a_j|m_j)\sum_{o_i',\omega_i'}T(s_f,a_i, a_j,s_f')\nonumber\\
&\times b_i(\langle s_f,s_r\rangle)Z(a_i, a_j, s_f, s_f', o_f')W_i(a_i,a_j,\omega_i')\left . V(\tau(b_i,a_i,o_i',\omega_i',b_i'))\right ]
\end{align}

\subsection{Model Belief Update}

The belief update in equation ~\ref{eq:10} updates the belief over states and other agent's model simultaneously. The state belief update and model belief update can be separated in case when the two parts are not handled by a single network. In a  two-agent setting,  agent $j$'s  model set at  time step  $t$ is
denoted  as  $M_j$,  where  $M_j$  contains  a  finite  number  of
pre-defined models  $m_j$. Given agent $i$'s  action $a_i$, public
observation $\langle o_f', o_r'\rangle$,  private observation $\omega_i'$,  and previous
belief $b_i$, the model belief update is defined as below:

\begin{align}
&b_i'(m_j'|\langle s_f',s_r' \rangle,b_i,a_i, o_i', omega_i') = \frac{Pr(m_j', \omega_i'|\langle s_f',s_r' \rangle,a_i,o_i',b_i)}{Pr(\omega_i'|\langle s_f',s_r' \rangle,,a_i, o_i', b_i)}\nonumber\\
&\propto\sum_{m_j}b_i(m_j) Pr(m_j', \omega_i'|\langle s_f',s_r' \rangle,a_i, o_i', m_j)\nonumber\\
&\propto\sum_{m_j}b_i(m_j) \sum_{a_j}Pr(m_j', w_i'|\langle s_f',s_r' \rangle,a_i, o_i', m_j, a_j)\nonumber\\
&\times Pr(a_j|\langle s_f',s_r' \rangle,a_i,o_i',m_j)\nonumber\\
%&\phantom{====}Pr(a_j^t|a_i^t,o^{t+1},m_j^t)\nonumber\\
&\propto\sum_{m_j}b_i(m_j) \sum_{a_j}Pr(a_j|m_j)Pr(m_j', \omega_i'|\langle s_f',s_r' \rangle,a_i,a_j, o_i',m_j)\nonumber\\
&\propto\sum_{m_j}b_i(m_j) \sum_{a_j}Pr(a_j|m_j)Pr(\omega_i'|m_j',\langle s_f',s_r' \rangle,a_i, a_j, o_i', m_j)\nonumber\\
&\phantom{====}Pr(m_j'|\langle s_f',s_r' \rangle,a_i,a_j, o_i',m_j)\nonumber\\
&\propto\sum_{m_j}b_i(m_j) \sum_{a_j}Pr(a_j|m_j) W_i(a_i,a_j,\omega_i')Pr(m_j'|a_i,a_j,o_i',m_j). \label{eq:1}
\end{align}
The last equation follows because the private observation function does not condition the private observation on the physical state. To simplify the term $Pr(m_j'|a, o_i',m_j)$, we substitute $m_j'$ with its components: $m_j' = (\pi_j', h_j')$, where $h_j'$ is agent $j$'s action-observation history at next time step and $\pi_j'$ is $j$'s policy.
\begin{align}
&Pr(m_j'|a_i,a_j, o_i',m_j)\nonumber\\
&=Pr(\pi_j', h_j'|a_i,a_j, o_i',\pi_j, h_j)\nonumber\\
&=Pr(h_j'|\pi_j',a_i,a_j,o_i', \pi_j, h_j)Pr(\pi_j'|\pi_j,a_i,a_j,o_i',h_j)\nonumber\\
&=Pr(h_j'|\pi_j',a_i,a_j,o_i', \pi_j, h_j)Pr(\pi_j'|\pi_j,a_i,a_j,h_j)\nonumber\\
&=\delta_K(APPEND(h_j, a_j, o_i'), h_j')\delta_K(\pi_j, \pi_j') \label{eq:2}
\end{align}
where $\pi_j^t$ is $j$'s policy contained in $m_j^t$ (note that the action-\\observation history in a model expands with time steps, but the policy in the model does not change). 

By substituting $Pr(m_j'|a_i,a_j,o_i,m_j)$ with equation~\ref{eq:2}, we can rewrite equation~\ref{eq:1} as 
\begin{small}
\begin{align}
&b_i'(m_j'|\langle s_f',s_r' \rangle,a_i, o_i', w_i', b_i)\propto \sum_{m_j}b_i(m_j)\sum_{a_j}Pr(a_j|m_j)\nonumber\\
&W_i(a_i,a_j,\omega_i') \delta_K(\pi_j,\pi_j')\sum_{o_i'}\delta_K(APPEND(h_j, a_j,o_i'),h_j')
\label{eq:model-bu}
\end{align}
\end{small}
The second Kronecker-delta function $\delta_K$ is 1 if the updated history matches the one in $m_j'$, it is 0 otherwise.

%-------------------------------------------------------------------------
\section{Interactive A2C$^+$}
\label{sec:iac-belief}
%-------------------------------------------------------------------------

Most current  deep reinforcement
learning  methods  require  explicit  exchange  of  information  among
agents.   Some  methods use  maximum  likelihood  estimation (MLE)  to
predict  other  agents'  actions from historical information in  scenarios  where  agents  cannot
exchange information. However, the result is unsatisfactory in complex
domains with large state and action spaces, as comparative evaluations
have  shown~\cite{maddpg}. In  this paper,  we present IA2C$^+$, which extends advantage actor-critic by maintaining predictions of other agents' actions based on dynamic beliefs over models. We implement A2C with one hidden layer with $tanh$ activation for input processing. For simplicity, we assume that both the critic and the actor networks map observations rather than beliefs, and that the critic maps observations to {\em joint action values}, $Q(\langle o_f,o_r\rangle,a_i,a_j)$. We estimate equation~\ref{eqn:advantage} as:
\begin{small}
\begin{align*}
  A(\langle o_f,o_r\rangle,a_i,\hat{a}_j) = avg[r + \gamma Q(\langle o_f', o_r'\rangle,a_i',\hat{a}_j') - Q(\langle o_f,o_r\rangle,a_i,\hat{a}j)]
 \end{align*}
 \end{small}
 while the actor's gradient (equation~\ref{eqn:gradient}) is estimated as:
\begin{small}
$$avg[\nabla_\theta\log\pi_\theta(a_i|\langle o_f,o_r\rangle)A(\langle o_f,o_r\rangle,a_i,\hat{a}_j)]$$
\end{small}
where $r,\langle o_f',o_r'\rangle$ and $a_i'$ are samples, $\hat{a}_j$ and $\hat{a}_j'$ are predicted actions (see next section), and the $avg$ is taken over sampled trajectories.

\subsection{Belief Filter}

We implement a Bayesian belief filter, integrated with the critic within the deep RL pipeline, to complete the model belief update. We further decompose the model belief update of equation~\ref{eq:model-bu} into two steps. The first step is the prediction, which accounts for other agent's actions. 
\begin{small}
\begin{align*}
\hat{b}_i'(m_j'&) = \sum_{m_j}b_i(m_j) \sum_{a_j}Pr(a_j|m_j) \delta_K(\pi_j,\pi_j') \delta_K(APPEND(h_j, o_i', a_j), h_j')
\end{align*}
\end{small}
Second step corrects the predictions using perceived observations. 
\begin{align*}
b_i'(m_j') \propto\hat{b}_i'(m_j')\sum_{a_j} W_i(w_i',a_i,a_j) 
\end{align*}

\vspace{-0.1in}
\begin{figure}[!ht]
\includegraphics[width=\columnwidth, height = 2.5in]{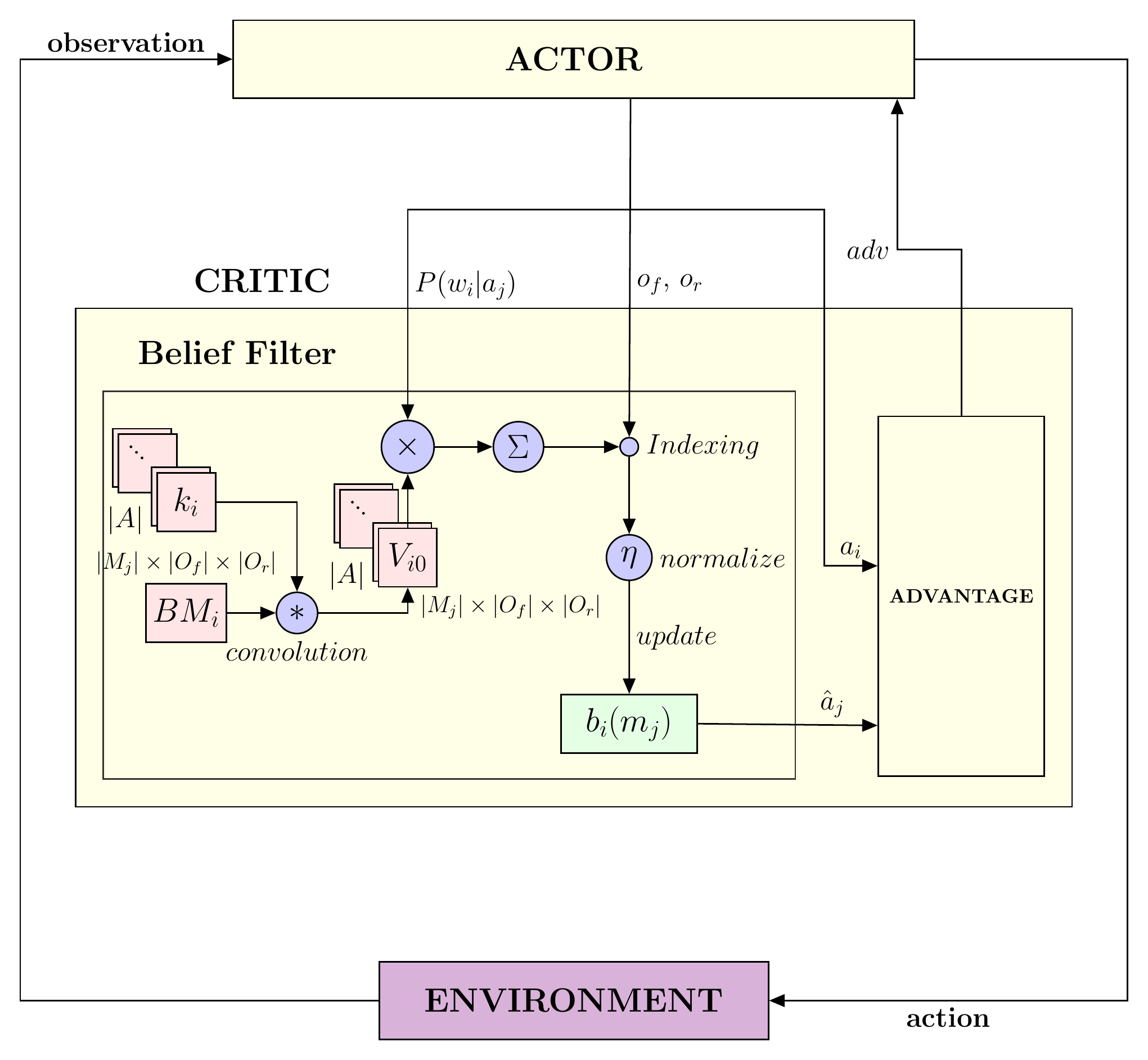}
\caption{\small Belief  filter  samples agent  $j$'s  action  based on  model
  distribution,  and  passes  the  predicted  action  to  agent  $i$'s
  interactive A2C network.}
\label{fig:IAC}
\end{figure}

Note that at each time  step, the belief is represented by  a $|M_j|  \times |O_f| \times |O_r|$  tensor. To keep the dimension of the belief matrix finite, we round $O_r$ to one decimal place. The  prediction step  is done by performing  convolution  on  belief   tensors  using  $|A|$-many  convolution
filters.  After   the  convolution,  we  obtain   $|A|$  belief  vectors
corresponding to  each possible joint action. Then  the correction  step is
executed by  multiplying each  belief vector  with the  probability of
perceiving its  corresponding private observation. Finally,  we sum up
all  result   vectors  and  index   all  possible  models   by  public
observations. This  step represents  the Kronecker-delta  function. We
use model leveraging to sample a predicted action of the other agent.

The whole workflow  of agent $i$'s A2C with belief  filter is shown in
Figure~\ref{fig:IAC}. Other agents have a similar network architecture.

%-------------------------------------------------------------------------
\section{Experiments}
\label{sec:experiments}
%-------------------------------------------------------------------------

We evaluate the  performance of the A2C method  with belief filtering,
labeled  as IA2C$^+$, on the cooperative-competitive {\sf Organization} domain using four sets of experiments.
First, we  show  that  a  method  that  does  not  account  for  the
history-dependent rewards fails to reach optimality. Second, we evaluate the need for cooperation in {\sf Organization} and that agents using the IA2C$^+$ can learn to cooperate. Third,  we explore  the  performances of  other  recent deep  MARL methods  on the  Organization domain  and  compare them  with IA2C$^+$. Finally,  we explore the  impact of  increasing noise levels in  the observations on the convergence  to the optimal policy by the various methods.

In order to model the partial observability of the problem, we extend the public observation $\langle o_f,o_r\rangle$ used in the indexing procedure to a short-term observation history that contains the current public observation as well as the previous public observation, i.e. $\{\langle o_f^{t-1},o_r^{t-1}\rangle, \langle o_f^t,o_r^t\rangle\}$. We include 5 models of the other agent in the pre-defined model set $M_j$. Three models lead to solely picking $self$, $balance$, and $group$ action, respectively, no matter which observation sequence is perceived. One model picks $group$ action for observation sequences $\{\langle o_e^{t-1},o_r^{t-1}\rangle, \langle o^t_e,o_r^t\rangle\}$, $\{\langle o^{t-1}_s,o_r^{t-1}\rangle, \langle o^t_e,o_r^t\rangle\}$, $balance$ action for observation sequence $\{\langle o^{t-1}_e,o_r^{t-1}\rangle,$
$\langle o^t_s,o_r^t\rangle\}$, $\{\langle o^{t-1}_s,o_r^{t-1}\rangle, \langle o^t_s,o_r^t\rangle\}$, and $self$ action for observation sequences $\{\langle o^{t-1}_m,o_r^{t-1}\rangle, \langle o^t_m,o_r^t\rangle\}$, $\{\langle o^{t-1}_m,o_r^{t-1}\rangle, \langle o^t_s,o_r^t\rangle\}$, $\{\langle o^{t-1}_s,o_r^{t-1}\rangle,$\\
$\langle o^t_m,o_r^t\rangle\}$. The last model picks $self$ action for observation sequences $\{\langle o^{t-1}_e,o_r^{t-1}\rangle,$
$\langle o^t_e,o_r^t\rangle\}$, $\{\langle o^{t-1}_s,o_r^{t-1}\rangle, \langle o^t_e,o_r^t\rangle\}$, $balance$ action for observation sequences $\{\langle o^{t-1}_e,o_r^{t-1}\rangle, \langle o^t_s,o_r^t\rangle\}$, $\{\langle o^{t-1}_s,o_r^{t-1}\rangle,$\\$ \langle o^t_s,o_r^t\rangle\}$, and $group$ action for observation sequences $\{\langle o^{t-1}_m,o_r^{t-1}\rangle,$\\$ \langle o^t_m,o_r^t\rangle\}$, $\{\langle o^{t-1}_m,o_r^{t-1}\rangle, \langle o^t_s, o_r^t\rangle\}$, and $\{\langle o^{t-1}_s, o_r^{t-1}\rangle, \langle o^t_m,o_r^t\rangle\}$.

\subsection{History-Dependent Rewards}
\label{a2cLSTM}

We introduce a baseline method, IA2C$^-$, which runs in the Organization domain that {\em does not} include the extra state and observation feature revealing the history-dependent reward. To compensate, it utilizes the LSTM for both actor and critic networks, in order to  model the dependence of its immediate rewards on the history of interactions. We  establish  the need  for utilizing  recurrence in this case, and  thereby  the need  to  correctly  model  the dependence on history, by   comparing its performance with IA2C$^-$ that  does not use LSTMs, using convolutional neural networks (CNN)  instead in the actor and critic.

\begin{figure}[!ht]
    \centering
    \includegraphics[width=.35\textwidth]{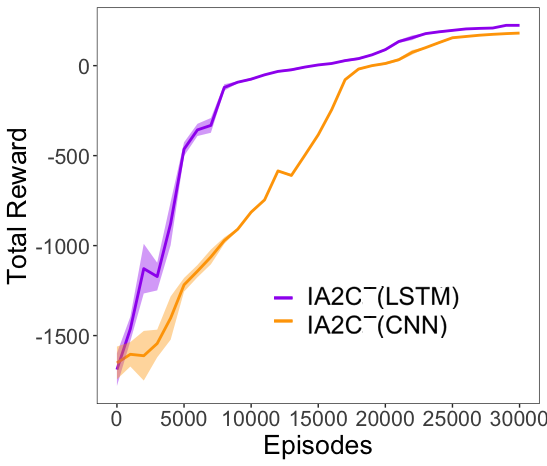}
    \caption{IA2C$^-$ without LSTM (uses a CNN instead) exhibits poor rewards for low numbers of episodes as we may expect, and eventually converges to a policy that is not optimal. }
    \label{fig:exp2}
\end{figure}

\begin{figure*}[!ht]
\begin{subfigure}{.33\textwidth}
    \centering
    \includegraphics[width=.95\textwidth]{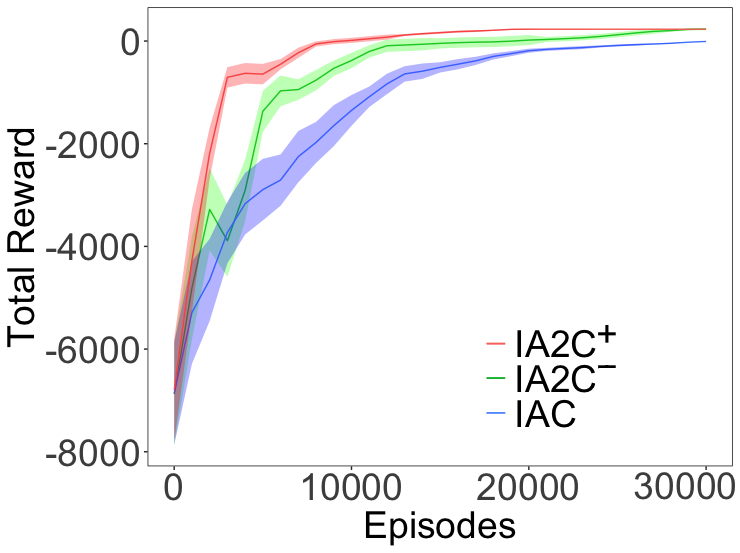}
    \caption{}
    \label{fig:4r}
\end{subfigure}
\begin{subfigure}{.33\textwidth}
    \centering
    \includegraphics[width=.95\textwidth]{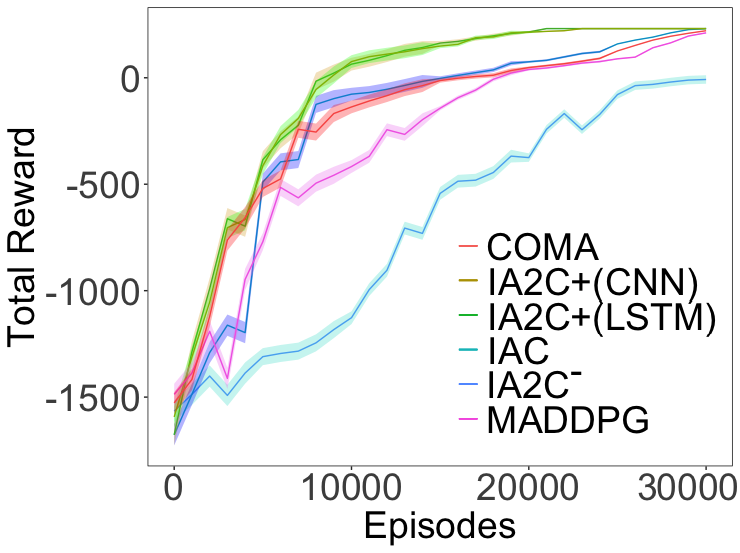}
    \caption{}
    \label{fig:exp3}
\end{subfigure}
\begin{subfigure}{.33\textwidth}
    \centering
    \includegraphics[width=.95\textwidth]{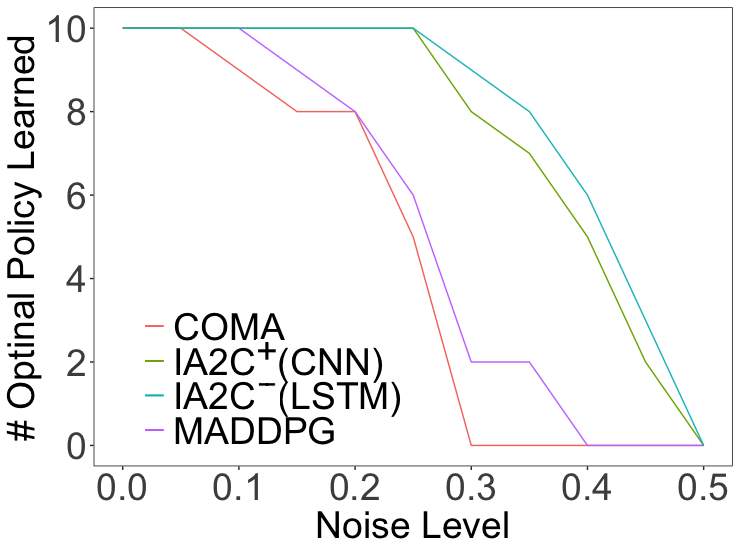}
    \caption{}
    \label{fig:exp5}
\end{subfigure}
\caption{\small (a) Four agents, each utilizing IAC or IA2C$^+$, converges (value loss becomes zero) to the optimal policy whereas IAC converges but not to the optimal policy. (b) IA2C$^+$ requires fewer episodes to converge to the optimal policy compared to MADDPG, COMA, and IAC. (c) Number of runs (out of 10) in which the optimal policy was learned as a function of the noise levels in private observations. IA2C$^+$ shows the most robust performance among the MARL techniques allowing for noise up to 0.3 while still learning the optimal policy.}
\vspace{-0.05in}
\end{figure*}

Notice  from  Figure~\ref{fig:exp2}  that   the IA2C$^-$ without  LSTM
converges, but not to the optimal policy in contrast to IA2C$^-$ with
LSTM that converges to the  optimal policy.  From this observation, we conclude that our approach to accommodation of history in Section~\ref{sec:ipomdp-model} is {\em not only sufficient but also necessary}, because if an alternative I-POMDP model existed that exhibited Markovian dynamics without requiring the extra features to enable optimal decisions, then IA2C$^-$ without LSTM should have reached the optimal policy as well.
We will use IA2C$^-$ with LSTM as a baseline method to compare with IA2C$^+$.

\subsection{Cooperation in the Organization Domain}

The optimal policy for the Organization domain involves performing the
$group$  joint action  in  response to  observation history that pertain  to
states  $s_{vl}$, $s_l$,  and  $s_m$; the  $balance$  joint action  in
response to observations  that pertain to state $s_h$;  and the $self$
action   in   response  to   observations   that   pertain  to   state
$s_{vh}$. Notice  that this involves  cooperation among the  agents at
various states. To quantify this, we  compare the value of the optimal
policy  with  that of  performing  the  $self$  joint action  for  all
observations  and the  $balance$  joint action  for all  observations.
Table~\ref{table:values}  shows a  significant difference  between the
three   values   indicating  that   both   cooperation   as  well   as
self-interestedness play a role in this domain.

\begin{table}[h!]
\centering
\begin{tabular}{ |c|c|c| } 
\hline
Optimal & Only Group & Only Balance \\
\hline
226.9 & 132 & 198 \\ 
\hline
\end{tabular}
\caption{\small Values of the optimal policy and other default behaviors for two agents.}
\label{table:values}
\vspace{-0.1in}
\end{table}

We hypothesize that independently learning  agents may not converge to
the cooperation  needed in this  domain. Consequently, we  compare the
performance of agents utilizing IA2C$^+$ with CNN as the neural net architecture with those utilizing IA2C$^-$ and
independent actor-critic (IAC) without belief filter  on   {\sf Organization} with  {\bf four}  learning  agents.
Figure~\ref{fig:4r} shows that agents learning using IA2C$^+$ and IA2C$^-$
learn the optimal policy within 30,000 episodes. We show the mean and standard deviation of 5 runs of each method in this chart. Four agents learn to unanimously pick  the $group$ individual action to  raise the organization's financial  health  level.   When  the  organization  is  in  a  better
financial health level, the four  agents coordinated to pick two $group$
actions, one  $balance$ action,  and one $self$ action in  order to
maximize the total  reward. However, IA2C$^+$ converges faster than IA2C$^-$ with the former requiring 20,000 episodes compared to the 30,000 episodes needed by the latter. It takes IA2C$^+$ about 3 hours to converge on a standard Linux 2.3GHz quad-core i7 processor with 8GB memory, while IA2C$^-$ takes about 10 hours to converge on the same machine. On  the other hand,  plain independent actor-critic  based learning failed to learn the optimal policy.

\subsection{Comparison with MARL Techniques}

Next, we explore the performance of previous state-of-the-art MARL methods such as MADDPG~\cite{maddpg} and COMA~\cite{coma} on the two-agent {\sf Organization} domain. We also include the results of IA2C$^-$ mentioned in Section~\ref{a2cLSTM} and a variant of IA2C$^+$ with LSTM. 
We disabled the policy exchange among agents as performed by MADDPG; instead each agent receives a noisy action whose noise probabilities are the same as those of the private observations in the Organization domain. As such, this simulates the effect of private observations, and makes it directly comparable to IA2C$^+$. As COMA employs a centralized critic, we noised the action sent by each actor to the critic to simulate the private observations.

We show the results in Figure~\ref{fig:exp3}. We point out that all methods eventually converge as their respective value losses become zero. While IA2C$^+$ with CNN and IA2C$^+$ with LSTM converges to the optimal policy at 20,000 episodes, and IA2C$^-$ converge to the optimal policy at 30,000 episodes, both MADDPG and COMA do not converge within these many episodes, instead requiring almost twice as many episodes. IA2C$^+$ with CNN has similar performance with IA2C$^+$ with LSTM while only using half the time to train. Closer inspection revealed that the belief filtering often predicted the true action of the other agent with a high probability despite the noisy observations. This more accurate prediction of the other agent's actions in about 88.6\% of the episodes, {\em despite none of the models in $M_j$ being  individually correct}, results in faster convergence to the optimal policy for each agent. IAC without belief filter shows the worst performance as we may expect, and fails to converge to the optimal policy as seen in the previous subsection.  

\subsection{Varying Private Observation Noise}

Finally, we vary the noisiness of the private observations to test the robustness of the methods to increasing uncertainty. We gradually increase the private observation noise level from 0 to 0.5, where noise level 0 means that the other agent's actions are perfectly observed, and noise level 0.5 means that there is only a 0.5 probability that the received observations indicate the correct actions of the other agent. 

We record the number of runs out of 10 for which each method learned the optimal policy for various noise levels. Figure~\ref{fig:exp5} shows the result of this experiment. IA2C$^+$ is able to consistently learn the  optimal policy when the private observation noise level is increased up to 0.3. Between noise levels 0.3 and 0.5, we observe an increasing number of runs where it fails to learn the optimal policy, failing completely beyond noise level 0.5. IA2C$^+$ with LSTM has a slightly higher performance than IA2C$^+$ with CNN, however, it requires almost twice as many episodes as IA2C$^+$ with CNN to converge. In contrast, COMA and MADDPG start to fail from noise level of around 0.1, and fail completely beyond noise levels 0.35 and 0.4, respectively. We conclude that IA2C$^+$ demonstrates consistent learning and robustness to higher levels of noise compared to the baselines in the {\sf Organization} domain.

%-------------------------------------------------------------------------
\section{Related Work}
\label{sec:related}
%-------------------------------------------------------------------------

While multi-agent RL mainly addresses purely cooperative or competitive tasks, there has been some work recently that address a mix of the two settings. We first discuss one prominent integrative work, the cooperative-competitive process (CCP), and then discuss recent work in MARL more generally, relating them to our contributions.

\subsection{Organization Domain as a CCP}
\label{subsec:ccp}

The CCP~\cite{ccp} is a framework for modeling mixed cooperative-competitive sequential decision problems. It blends both cooperative and competitive multi-agent modeling by introducing a slack parameter which controls the amount of cooperation versus competition. A group-dominant CCP (GD-CCP) first maximizes the group reward, then optimizes the individual reward while allowing a deviation of up to the slack from the optimal group reward. Individual-dominant CCP (ID-CCP) follows a similar pattern but with a reversed preference. Non-linear programming (NLP) is used for solving the CCPs. Our work can be seen as a pragmatic generalization of the CCP, situated in a MARL context, as discussed in the next paragraph. Instead of manually defining a parameter to choose between cooperate or compete, Kleiman-Weiner et al.~\cite{hierarchical} present a hierarchical model that integrates low-level action plans to high-level strategy. An agent can infer other agents' high level strategy before deciding whether to select a cooperative or a competitive strategy, and thus perform corresponding low-level actions specified by selected strategies. 

Because  the CCP as is cannot model  the organization
domain, we generalize it to a history-dependent CCP by adding $R_{-1}$
to the  reward vector, indicating  the history-dependent  component of
the rewards.  In  addition, we further decompose  the observation into
public and  private. Public observations  depend on the state  and the
joint  action,  while private  observations  depend  on other  agent's
individual actions. Also, the individual rewards $R_i$ depend on agent
$i$'s  individual action  only, instead  of  the joint  action in  the
original CCP definition. The history-dependent CCP for {\sf Organization} is given below.
\begin{itemize}[leftmargin=*, itemsep=0in]
\item $I = \{1,2,\dots,n\}$ is the set of $n$ agents
\item $S = S_f\times S_r$, where $S_f=\{s_{vl}, s_l, s_m, s_h, s_{vh}\}$ and $S_r$ is the continuous state feature for memorizing the previous reward
\item $\vec{A} = \{self, balance,  group\}$ is the set of 3
  joint actions, determined by the majority choices of the $n$ agents,
  as given in Section~\ref{subsec:actions}
\item $T:S  \times \vec{A}  \times S \rightarrow  [0,1]$ is  the state
  transition function mapping state $s$  and joint action $\vec{a}$ to
  successor state $s'$, such that $T(s,\vec{a},s') = Pr(s'|s,\vec{a})$
\item $O = \{o_e, o_s, o_m\}$ is the set of public observations
\item $Z:\vec{A}  \times S \times  O \rightarrow [0,1]$ is  the public
  observation function  mapping joint  action $\vec{a}$  and successor
  state    $s'$     to    a     public    observation     such    that
  $Z(\vec{a}, s',o) = Pr(o|\vec{a},s')$
\item  $\Omega  =  \{\omega_s,\omega_b,\omega_g\}$  is the  set  of  3
  private observations of each agent
\item  $W:  A   \times  \Omega  \rightarrow  [0,1]$   is  the  private
  observation function  mapping individual  action $a_j$ to  a private
  observation             $\omega_i$             such             that
  $W(a_j,\omega_i) = Pr(\omega_i|a_j)$
\item $\vec{R} = [R_{-1}, R_0, R_1, R_2, \dots, R_n]$ is the vector of
  rewards; for each state $s$,  $R_0(s,\vec{a})$ is a group reward for
  all agents and  $R_i(s,a_i)$ is an individual reward  for each agent
  $i$. $R_{-1}$ is  the discounted history-dependent  component of the
  reward, $R_{-1}(\langle s_f,s_r\rangle)=\phi\cdot s_r$.
\end{itemize}
When  $\phi=0$ and $\Omega$ is empty,  the  history-dependent  CCP
reduces to the original CCP. While Wray et al.~\cite{ccp} solve the original CCP in a centralized manner for all agents, our approach is analogous to solving this variant from each individual agent's perspective.

\subsection{Multi-agent RL}

Multi-agent deep deterministic policy gradient (MADDPG)~\cite{maddpg} is a multi-agent RL algorithm that can be applied to mixed cooperative-competitive settings. It extends actor-critic by exchanging policies among agents' critic, while the actor only has access to local information. After training is completed, only the actors are deployed in the environment. When direct policy exchange is not possible, each agent maintains an approximation to the true policies of the other agents. The approximation is done by maximizing the log likelihood of the other agents' actions (which the learner is able to observe). However, when the other agents' actions are not perfectly observed due to noise, MADDPG is unable to learn the optimal policy, as our experiments on {\sf Organization} have shown. 

Learning with opponent-learning awareness (LOLA)~\cite{lola} is another actor-critic method where the agents attempt to directly influence the policy updates of other agents. Instead of learning the best response, LOLA learns to maximize the expected return after the opponent updates its policy with one naive learning step. In this way, a LOLA learner explicitly accounts for the learning of other agents in the environment, within its own learning. LOLA is suitable for both cooperative and competitive problems. Nevertheless, LOLA requires that the agents have access to each others' exact gradients. LOLA with opponent modeling (LOLA-OM) removes this limitation of accessing other agent's policy gradients, by estimating the other agent's gradients from the trajectories, using maximum likelihood estimation. However, it is not straightforward to accommodate either the exact gradients or the estimated ones as private observations, in a manner consistent with the private observations of IA2C$^+$, or those of our modified MADDPG and COMA for {\sf Organization}, thus precluding a fair comparison with these methods in our experiments. For this reason, we have excluded LOLA from our set of baselines.

Independent deep Q-Learning (IQL)~\cite{iql} extend deep Q-Learning network (DQN)~\cite{mks15} architecture to multi-agent settings by allowing the agents to select actions independently, and to receive separated individual rewards from the environment. IQL can learn policies ranging from fully cooperative to competitive by tuning the reward function. It demonstrates the possibility of decentralized learning in complex multi-agent environments. As we have already included IAC in our set of baselines, and Q-learning is an  off-policy technique compared to the on-policy actor-critic based algorithms used in our experiments, we exclude IQL from our set of baselines.

Counterfactual multi-agent policy gradients (COMA)~\cite{coma} is also an actor-critic method which consists of a centralized critic and multiple (decentralized) actors. COMA uses a centralized critic to compute the agent-specific advantage functions that compares the estimated return for the current joint action to a counterfactual baseline that marginalizes out one single agent's action at a time, while keeping all other agents' actions fixed. However, COMA requires access to the true state or the joint action-observation history.

%-------------------------------------------------------------------------
\section{Concluding Remarks}
%-------------------------------------------------------------------------

We introduced the {\sf Organization} domain, inspired by typical real-world businesses, where agents must both cooperate and compete to attain optimal behavior. Agents in this domain not only receive noisy observations about the state and others' actions, but also obtain rewards that, in part, depend on the total reward of the previous time step, analogous to bonus pay. Subsequently, the Organization domain offers substantially more realistic challenges than previous MARL domains. The presence of history-dependent rewards challenges the applicability of traditional decision-making frameworks and the need for cooperation precludes independent learning in this domain. We presented a new method that combined decentralized actor-critic based learning with maintaining beliefs over a finite set of candidate models of the other agents. This method is comparatively robust to noisy observations and converges significantly faster to the optimal policy in the Organization domain compared to previous state-of-the-art MARL baselines. An immediate avenue of future work is to further scale the number of agents beyond four to better simulate real-world business organizations. 

%-------------------------------------------------------------------------
%\clearpage
\bibliographystyle{BST/ACM-Reference-Format}  % do not change this line!
\bibliography{hbdAAMAS21.bib}  % put name of your .bib file here
%-------------------------------------------------------------------------

\end{document}